\documentclass[conference]{IEEEtran}
\IEEEoverridecommandlockouts
\usepackage{cite}
\usepackage{amsmath,amssymb,amsfonts}
\usepackage{algorithmic}
\usepackage{graphicx}
\usepackage{textcomp}
\usepackage{xcolor}
\usepackage[most]{tcolorbox}
\usepackage[export]{adjustbox}
\usepackage{caption}
\usepackage{subcaption}\usepackage{array}
\usepackage{multirow}
\usepackage{booktabs}
\usepackage{eso-pic}

\AddToShipoutPictureBG*{%
  \AtPageUpperLeft{%
  \raisebox{-1cm}[0pt][0pt]{%
    \makebox[\paperwidth][c]{%
      \parbox{\textwidth}{\centering\color{gray!30}\small
        This work has been submitted to the IEEE for possible publication. Copyright may be transferred without notice, after \\
which this version may no longer be accessible
        }
        }
        }
  }%
}

\def\BibTeX{{\rm B\kern-.05em{\sc i\kern-.025em b}\kern-.08em
    T\kern-.1667em\lower.7ex\hbox{E}\kern-.125emX}}
\begin{document}

\title{GPT-4 for Occlusion Order Recovery}

\author{\IEEEauthorblockN{Kaziwa Saleh\IEEEauthorrefmark{1}\IEEEauthorrefmark{2}, Zhyar Rzgar K Rostam\IEEEauthorrefmark{1}\IEEEauthorrefmark{2}, S\'{a}ndor Sz\'{e}n\'{a}si\IEEEauthorrefmark{2}\IEEEauthorrefmark{3}, Zolt\'{a}n V\'{a}mossy\IEEEauthorrefmark{2}}  
\IEEEauthorblockA{\IEEEauthorrefmark{1}Doctoral School of Applied Informatics and Applied Mathematics, Obuda University, Budapest, Hungary \\}  
\IEEEauthorblockA{\IEEEauthorrefmark{2}John von Neumann Faculty of Informatics, Obuda University, Budapest, Hungary \\}
\IEEEauthorblockA{\IEEEauthorrefmark{3}Faculty of Economics and Informatics, J. Selye University, Komárno, Slovakia\\}
Emails: \{kaziwa.saleh, kwekha.rostam.zhyar, szenasi.sandor, vamossy.zoltan\}@nik.uni-obuda.hu
}

\maketitle

\begin{abstract}
Occlusion remains a significant challenge for current vision models to robustly interpret complex and dense real-world images and scenes. To address this limitation and to enable accurate prediction of the occlusion order relationship between objects, we propose leveraging the advanced capability of a pre-trained GPT-4 model to deduce the order. By providing a specifically designed prompt along with the input image, GPT-4 can analyze the image and generate order predictions. The response can then be parsed to construct an occlusion matrix which can be utilized in assisting with other occlusion handling tasks and image understanding. We report the results of evaluating the model on COCOA and InstaOrder datasets. The results show that by using semantic context, visual patterns, and commonsense knowledge, the model can produce more accurate order predictions. Unlike baseline methods, the model can reason about occlusion relationships in a zero-shot fashion, which requires no annotated training data and can easily be integrated into occlusion handling frameworks.

\end{abstract}


\section{Introduction}
Machines are demonstrating increasing efficacy in perceiving and interpreting their surrounding environments. Achieving this presents significant challenges, particularly when considering the inherent complexity of real-world scenes. A key difficulty arises from the predominance of partially visible objects compared to fully visible ones \cite{ao2023image}. Consequently, the accurate recognition of objects and the inference of their ordinal relationships are fundamental prerequisites for robust image and scene understanding and various downstream applications.

However, order recovery is inherently challenging due to occlusion. Occlusion occurs where one object obstructs the view of another, and it happens in various forms, ratio, and position. Furthermore, an object may either occlude other objects or be occluded by one or more objects, further complicating the task of image understanding \cite{saleh2021occlusion}. Recently, several notable works have addressed occlusion, such as predicting the full gestalt of occluded instances \cite{ozguroglu2024pix2gestalt, xu2024amodal, zhan2024amodal, liu2025towards}, and completing their appearances \cite{ao2025open}. However, in terms of order recovery and occlusion detection, almost all works in the literature \cite{zhu2017semantic, zhan2020self, lee2022instance} depend on retrieving the amodal mask (the segmentation mask of the object including its occluded region) of the objects to predict the occlusion relationship between them. This requires training the model on annotated occluded dataset which is not as commonly available as non-occluded ones.

Large language models (LLMs) posses impressive abilities in text generation, contextual learning and reasoning. Their reasoning and parsing capabilities can be leveraged to interpret visual content for vision-centric tasks such as detection, visual grounding, instance segmentation, and image captioning \cite{wang2023visionllm}. Additionally, recent versions of GPT, specifically GPT-4 \cite{achiam2023gpt} have opened new avenues for visual reasoning tasks \cite{ray2023chatgpt}, including occlusion order recovery. 

In this work, we use a pre-trained GPT-4 model to infer the occlusion order of objects within an input image. We provide the model with both the image and a textual prompt that specifies the desired response format, preventing overly detailed outputs. To ensure the model focuses on specific instances and that detected object names align with ground truth labels, we include a list of relevant object categories found in the image. The generated output is then parsed to extract an occlusion matrix, which represents the ordinal relationships between the objects.

We tested the model on COCOA \cite{zhu2017semantic} and InstaOrder \cite{lee2022instance} datasets and results demonstrate that GPT-4 makes more accurate order predictions than baseline models. Our method is simple yet effective. Unlike other methods that rely heavily on geometric cues, segmentation masks, or supervised learning with annotated datasets, using a pre-trained LLM enables reasoning based on semantic context and commonsense knowledge. GPT-4 can infer likely occlusion relationships between objects without requiring pixel-level training. This approach allows the model to generalize across various scenes and object types. In contrast to prior works, our method leverages the pre-trained capabilities of GPT-4 to reason about occlusion in a zero-shot fashion, making it adaptable, and less dependent on extensive labeled data. To the best of our knowledge, this is the first work that addresses the problem of occlusion order recovery utilizing a pre-trained GPT-4 model. 

\section{Related Work}
To predict the order of the detected objects, Yang et al. \cite{yang2011layered} introduced a layered object detection and segmentation method. In \cite{tighe2014scene}, a semantic label for each pixel is determined and objects are subsequently ordered according to their inferred occlusion. From a single monocular image, Zhang et al. \cite{zhang2015monocular} create instance level segmentation and determine depth orderings using a convolutional neural network and a Markov random field. Zhu et al. \cite{zhu2017semantic} predict the depth relationship between object pairs. They do this using their manually annotated dataset, which includes both occlusion ordering and segmentation masks, along with a supervised model called OrderNet$^{M+1}$. This model needs an image and two masks to figure out which object occludes the other.
Also, Ehsani et al. \cite{ehsani2018segan}, reconstruct the amodal mask of an object and then deduce its depth order from how objects occlude each other. Similarly, the authors in \cite{zhan2020self} use a self-supervised model called PCNet-M to predict amodal masks for objects. They then infer the occlusion order by determining which of two overlapping objects requires more completion, identifying it as the one being occluded. Furthermore, Lee and Park \cite{lee2022instance} developed InstaOrder$^{o,d}$, a model that uses a pre-trained ResNet-50 \cite{he2016deep} and two fully connected layers to simultaneously predict both the occlusion and depth order from a pairwise segmentation mask and an image patch. The authors in \cite{li2022distance} propose a three-decoder architecture with a generalized intersection box prediction to pay more attention to relevant information in order to determine the order of occlusion and distance of objects. In contrast to the approaches mentioned above, Saleh and V\'{a}mossy \cite{saleh2022bbbd} introduced BBBD, an approach that determines the occlusion order of overlapping objects without any training. This is achieved by utilizing the bounding boxes and modal segmentation masks. Their method identifies the occluding object by finding the intersection area between two objects; the one with a larger mask within that specific region is considered the occluder.
In contrast to prior works, our method leverages the pre-trained capabilities of GPT-4 to reason about occlusion order in a zero-shot fashion.

\section{GPT-4}
AI and deep learning techniques, particularly Transformer-based models \cite{vaswani2017attention}, have recently achieved remarkable success across a wide range of tasks. GPT-4 is a powerful large language model (LLM) designed based on the Generative Pre-trained Transformer (GPT) architecture by OpenAI with diverse applications across various professional and academic domains \cite{ray2023chatgpt,biswas2023prospective, laki2023sentiment,wu2023visual}. It leverages the self-attention mechanism at the core of Transformer model to process the input sequence and generate outputs. This mechanism allows the model to focus on different parts of the input simultaneously.

Although OpenAI has not publicly disclosed the exact architectural details of GPT-4, it is substantially larger and more capable than its predecessors. Unlike earlier versions, certain variants of GPT-4 are multi-modal and can process and reason over images alongside textual inputs, enabling the use of linguistic reasoning approaches in vision processing \cite{wu2023visual}. One such task is object ordering recovery.

\section{Methodology}
To recover occlusion order from images using a pre-trained GPT-4 model, we supply a carefully designed textual prompt that constrains the format and content of the output. Without such constraints, the model tends to generate highly detailed scene descriptions, which are difficult to evaluate systematically. Therefore, to enable comparison with ground truth ordering, both pre-processing of the input data and post-processing of the model output are necessary.

For each image in the dataset, the category of objects is extracted and embedded into the prompt. The labels can be extracted using any pre-trained detection model. If there are multiple instances of the same category in the image, they are numbered (e.g. bottle 0, bottle 1, etc) to ensure uniqueness. Fig.~\ref{fig:prompt} presents an example of the provided prompt. The categories enforce the model to produce descriptions composed only of the given object categories. For each image, GPT-4 generated statements indicating pairwise occlusion relationships, such as "\textit{Object A occludes Object B}". These outputs are then programmatically parsed to construct an occlusion matrix encoding all detected pairwise object relationships. For each parsed statement, the object preceding "occludes" is designated as the occluding object, and the object following it is identified as the occluded object.

\begin{figure}[ht]
\centering
\begin{tcolorbox}[colback=gray!5, colframe=black,]
\textit{List all visible objects stated in these \{categories\} from foreground to background starting from index 0.\\
Professionally state object occlusion, return it in this format: \\
"Object A occludes Object B"\\
If there are multiple objects in same class, number them (e.g., bottle 0, bottle 1, etc).\\
Return only the ordered list and occlusions -- no explanations.}
\end{tcolorbox}
\caption{Example of the prompt sent to GPT-4 for occlusion order recovery. The categories are extracted from the image and passed as a CSV file.}
\label{fig:prompt}
\end{figure}

\begin{figure}[!h]
\centerline{\includegraphics[height=0.6\textwidth, width=0.3\textwidth]{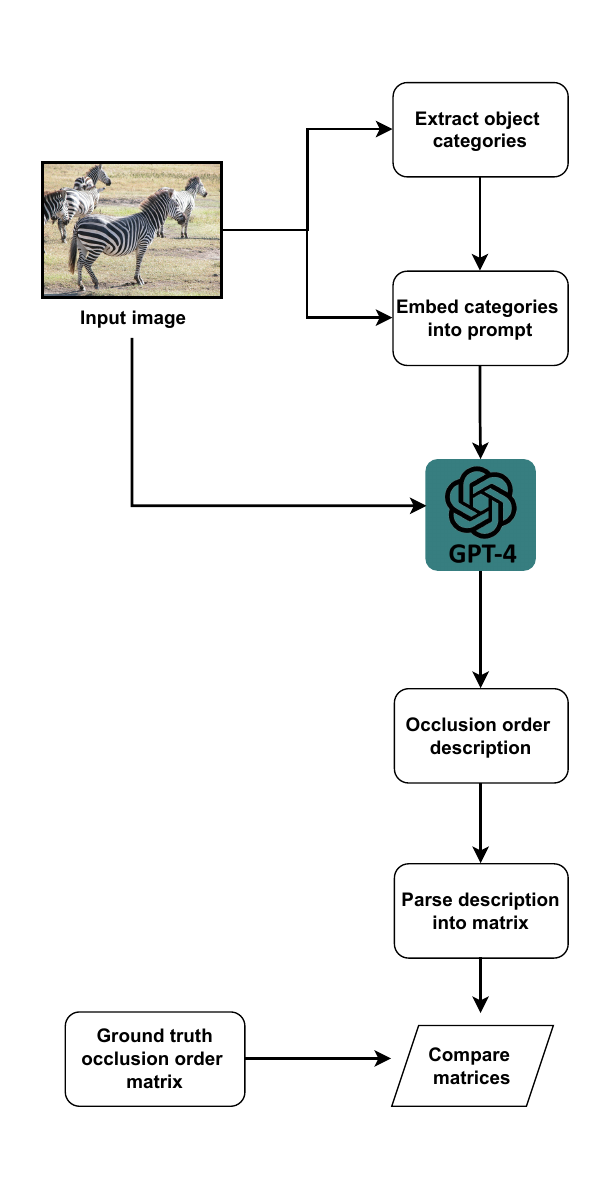}}
\caption{Workflow of the proposed approach}
\label{fig:model}
\end{figure}

Without supplying the explicit list of categories, the model's output may diverge from the ground truth annotations. For example, the model might refer to an object as a “car” or “automobile,” while the ground truth annotation labels it simply as a “vehicle”. By constraining the vocabulary, we ensure that occlusion relationships are defined consistently across all samples. Fig.~\ref{fig:model} illustrates the steps mentioned above.

\begin{figure*}[!h]
\centering
    
    \begin{subfigure}[t]{0.8\textwidth}
        \includegraphics[width=\textwidth,height=0.12\textheight]{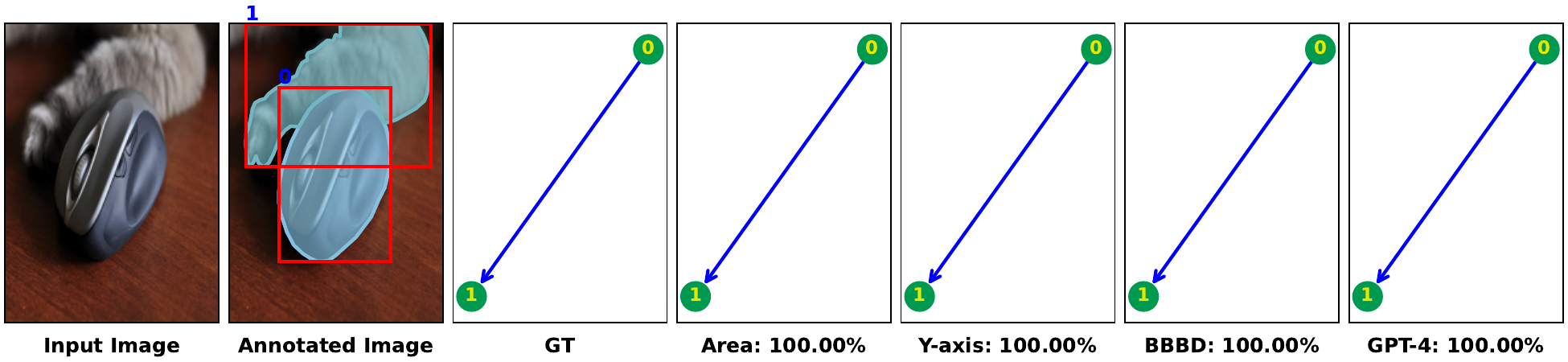}
        \caption{}
    \end{subfigure}
    
    \begin{subfigure}[t]{0.8\textwidth}
        \includegraphics[width=\textwidth,height=0.12\textheight]{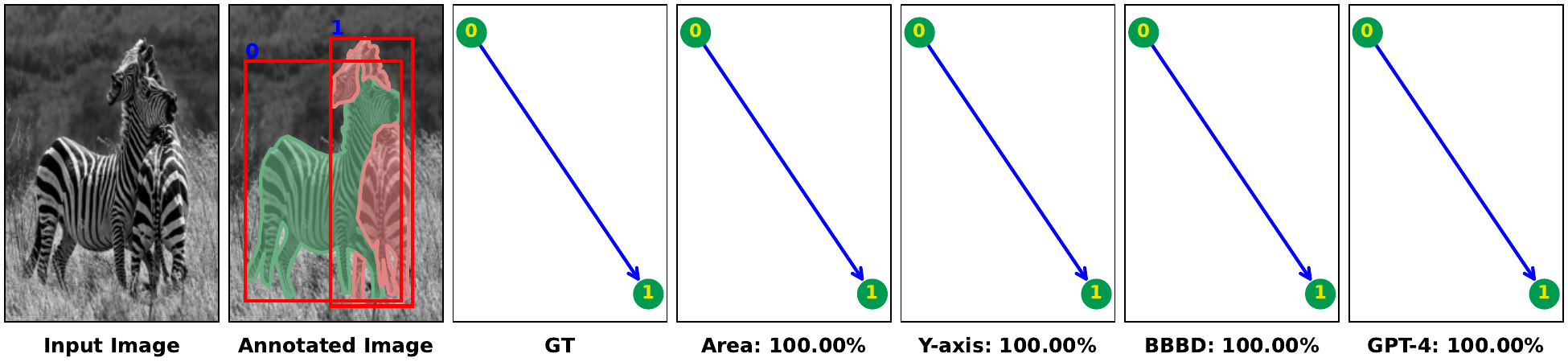}
        \caption{}
    \end{subfigure}
    \begin{subfigure}[t]{0.8\textwidth}
        \includegraphics[width=\textwidth,height=0.12\textheight]{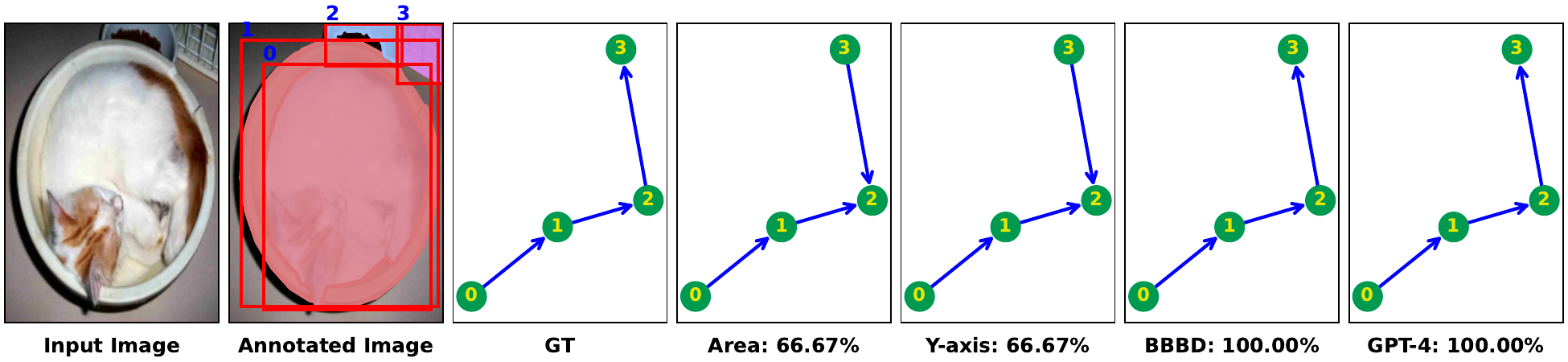}
        \caption{}
    \end{subfigure}
    \hfill
    \begin{subfigure}[t]{0.8\textwidth}
        \includegraphics[width=\textwidth,height=0.12\textheight]{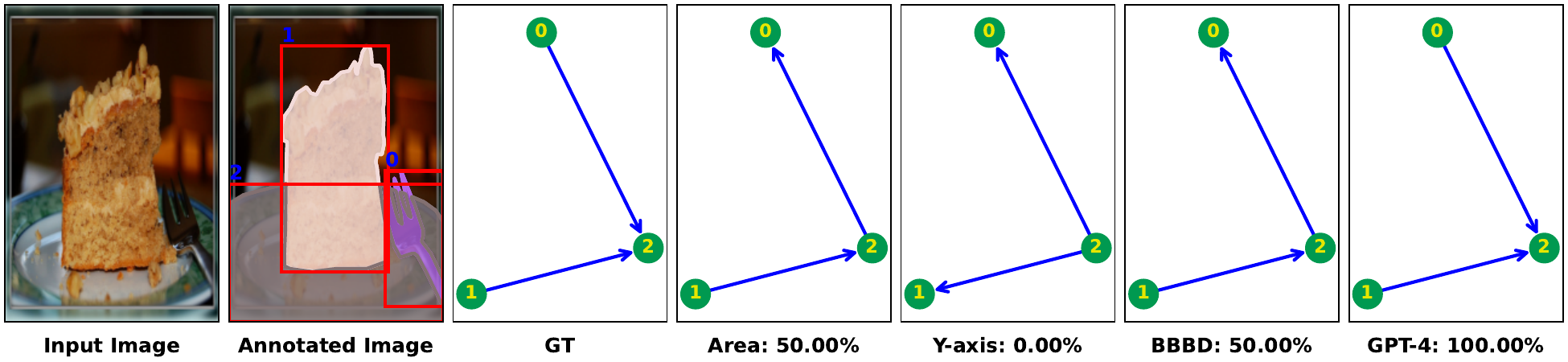}
        \caption{}
    \end{subfigure}
    \hfill
    
\caption{Samples demonstrate how a pre-trained GPT-4 model outperforms baselines. While it achieves comparable results to other methods in (a) and (b), samples in (c) and (d) clearly highlight the model’s effectiveness}
\label{fig:samples}
\end{figure*}

\section{Datasets}
\textbf{COCOA}: COCOA is a subset of COCO \cite{lin2014microsoft} dataset. It contains 2500 images in the training set with 22163 instances, and 1323 images in the validation set with 12753 instances. The dataset is annotated with amodal, modal segmentation masks, and pair-wise occlusion ordering. 

\textbf{InstaOrder}: InstaOrder is built on COCO 2017 dataset. It includes 2,859,919 instance-level occlusion and depth ordering of 503,939 instances from 100,623 images. All other metadata, such as object categories, bounding boxes, and segmentation masks were sourced directly from the original COCO annotations. Therefore, the two annotation sets were merged by matching their records based on the shared image identifiers.

Since the language model is pre-trained, we only evaluate it on the validation set. The category (or the name) of instances are extracted and used in the prompt that is given to GPT-4.

\section{Results and Discussion}
To assess the accuracy of our method, we conduct a comparative evaluation against three baseline approaches: Area, which assumes that the larger object is the occluder; Y-Axis, which considers the object located lower along the vertical axis as the occluder; and BBBD\cite{saleh2022bbbd}.

The occlusion order matrices predicted by the pre-trained GPT-4 model are compared with the ground truth annotations. When multiple occluded objects are present in the scene, for each pair of objects (\textit{i}, \textit{j}), the model predicts whether object \textit{i} occludes object \textit{j}. The accuracy is then computed as the proportion of correctly predicted pairwise relationships compared to the ground truth occlusion matrix.

Table~\ref{table:results} reports the accuracy scores obtained by our method and the baselines across the evaluation datasets. On COCOA, the pre-trained GPT-4 approach achieved improvements of 15\%, 20\%, and 12\% in accuracy over the Y-Axis, Area, and BBBD baselines, respectively. On the InstaOrder dataset, the model demonstrated further increase in accuracy, achieving 20\%, 11\%, and 26\% higher accuracy compared to Y-Axis, Area, and BBBD. These results indicate that leveraging a pre-trained large language model provides a substantial advantage in recovering occlusion order over conventional heuristic baselines. Fig.~\ref{fig:samples} illustrates the effectiveness of GPT-4 in order prediction through an order graph. In this graph, nodes represent individual objects, and directed edges indicate occlusion relationship, where the object at the tail of an edge is the occluder.

\begin{table}[!h]
\caption{Accuracy results for occlusion order recovery.}
\label{table:results}
\centering
    \begin{tabular}{ p{0.2\linewidth} p{0.13\linewidth} p{0.13\linewidth} p{0.13\linewidth} p{0.13\linewidth} } \hline  
     & Area & Y-axis & BBBD & GPT-4 \\ 
     \hline
        \hline
        \textbf{COCOA} & 65.43\% & 61.36\% & 69.53\% & \textbf{82.26\%} \\
        \textbf{InstaOrder} & 52.23\% & 62.02\% & 47.72\% & \textbf{73.05\%}\\ 
       \hline \hline
    \end{tabular}
\end{table}

\begin{figure*}[!h]
\centering
    \begin{subfigure}[t]{0.8\textwidth}
        \includegraphics[width=\textwidth,height=0.12\textheight]{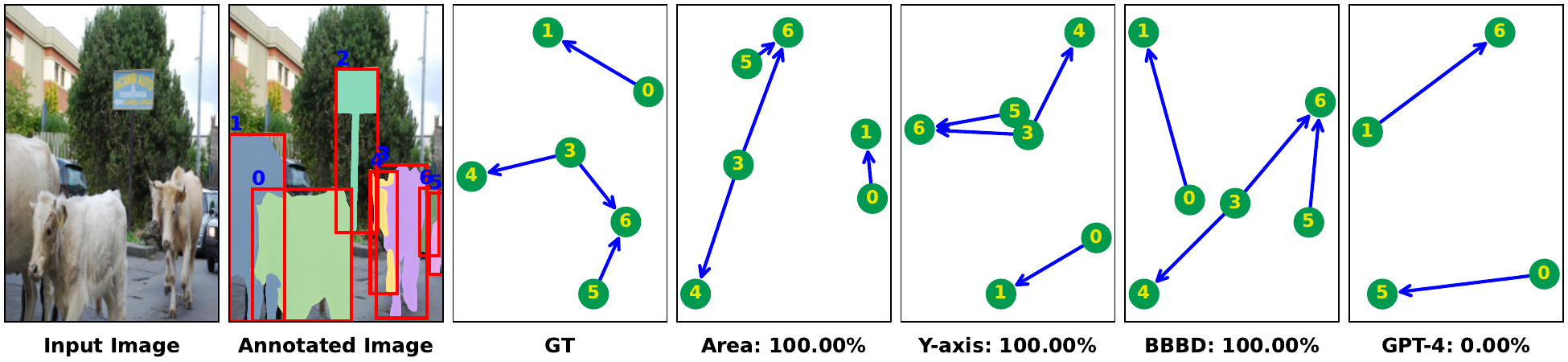}
        \caption{}
    \end{subfigure}
    \hfill
    \begin{subfigure}[t]{0.8\textwidth}
        \includegraphics[width=\textwidth,height=0.12\textheight]{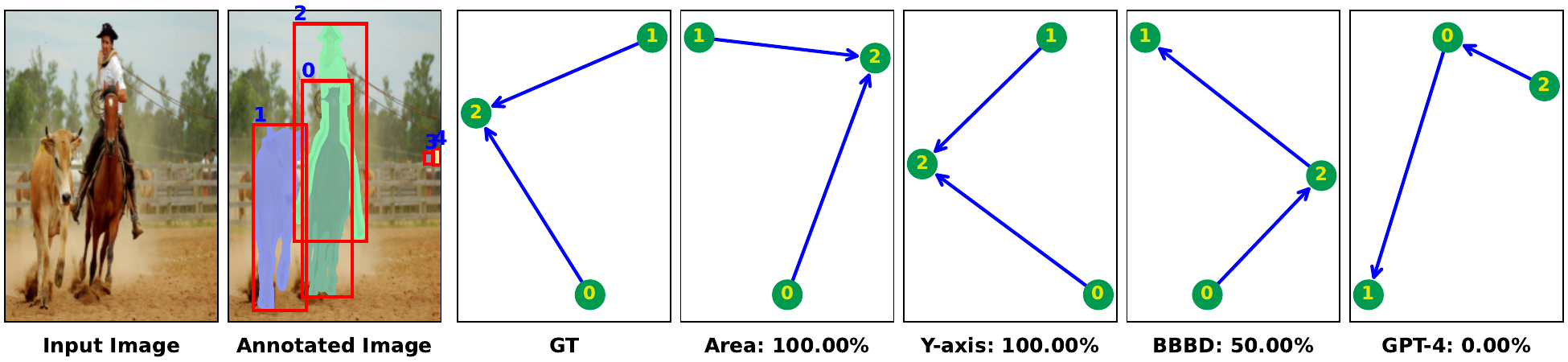}
        \caption{}
    \end{subfigure}
    \hfill
    \begin{subfigure}[t]{0.8\textwidth}
        \includegraphics[width=\textwidth,height=0.12\textheight]{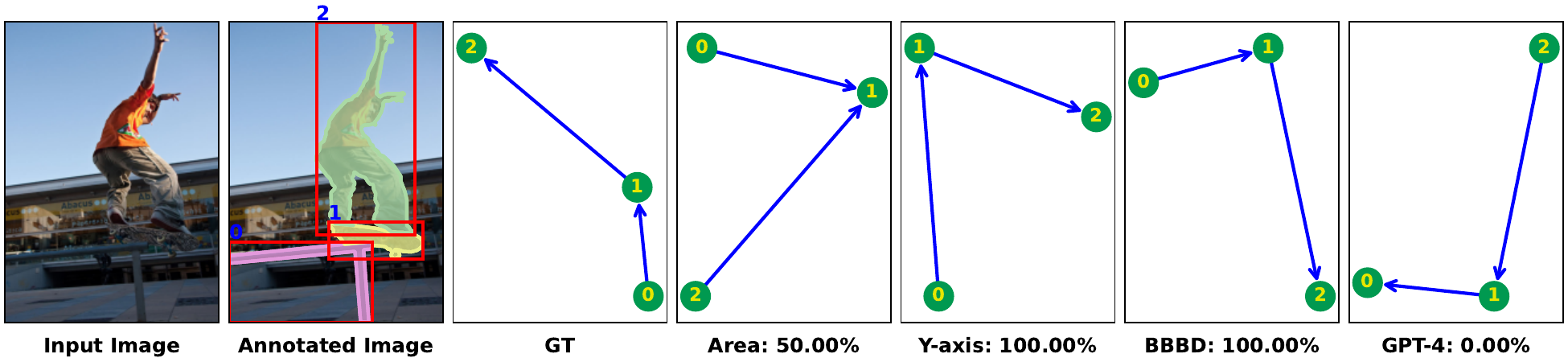}
        \caption{}
    \end{subfigure}
    
    \begin{subfigure}[t]{0.8\textwidth}
        \includegraphics[width=\textwidth,height=0.12\textheight]{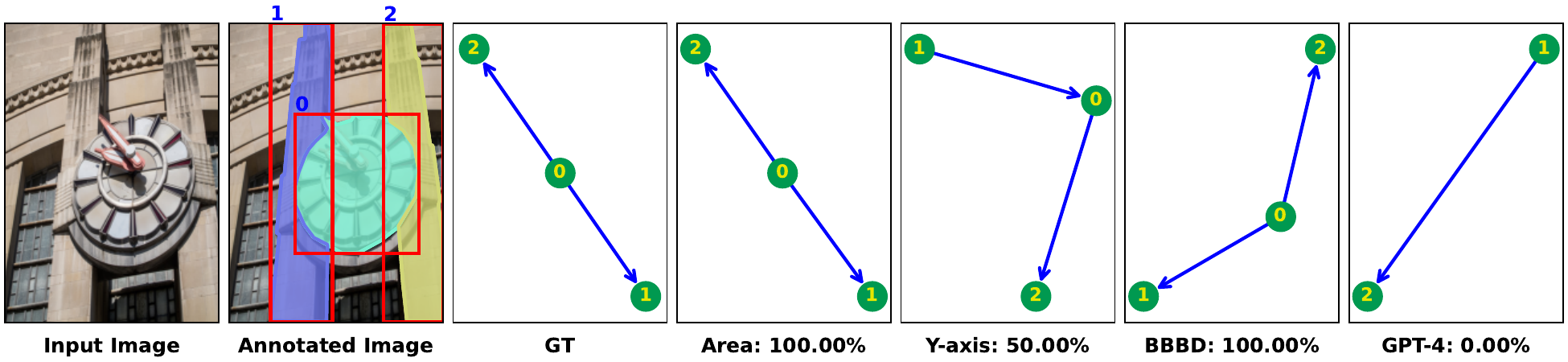}
        \caption{}
    \end{subfigure}

        \begin{subfigure}[t]{0.8\textwidth}
        \includegraphics[width=\textwidth,height=0.12\textheight]{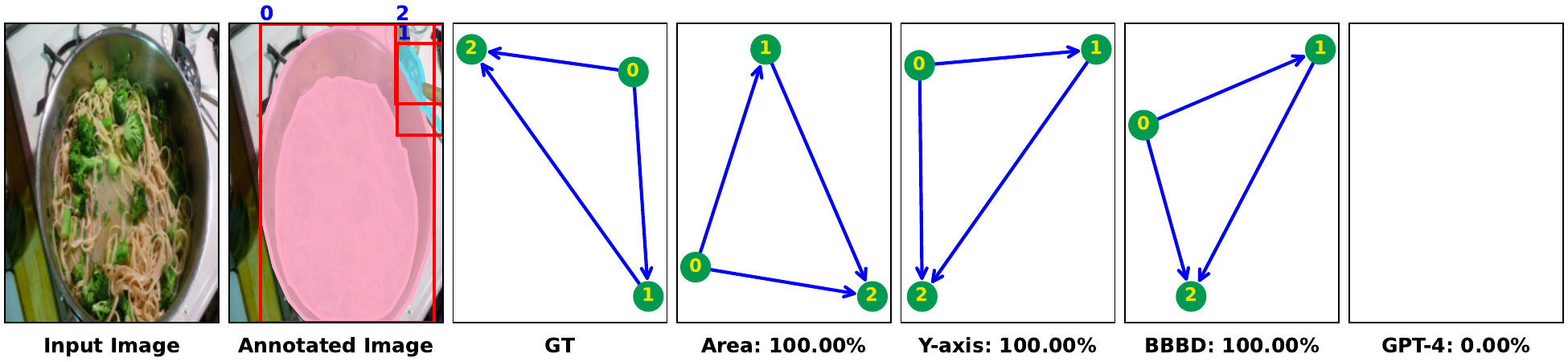}
        \caption{}
    \end{subfigure}
\caption{Examples of failure cases where the pre-trained GPT-4 model does not produce correct predictions.}
\label{fig:failureSamples}
\end{figure*}

However, as shown in the table, the model produces incorrect predictions in 17.74\% of the COCOA cases and 26.95\% of the InstaOrder cases. Examples of such failures are illustrated in samples (a)–(d) in Fig.~\ref{fig:failureSamples}. Among these, 7.71\% of the COCOA samples and 7.05\% of InstaOrder samples include cases where the model fails to predict any occlusions, resulting in an all-zero matrix, as depicted in sample (e) of the same figure. The reason for these failures can be attributed to several reasons: ambiguous overlapping of objects, where the model struggled to distinguish between closely positioned items; a mismatch in categories, where the labels predicted by the model did not align with the ground truth; or a sequence mismatch, where the order in which the model detected objects differed from the ground truth sequence.

Furthermore, when the list of object categories is not given in the prompt, GPT-4 tends to identify a greater number of objects, as illustrated in Table~\ref{table:no_category}. While this demonstrates the model’s capacity to predict additional instances along with their corresponding occlusion relationships, it also complicates the evaluation process against the ground truth annotations. Specifically, the difference in the number of detected objects makes accuracy computation infeasible or, in some cases, entirely impractical.

Despite these issues, the results still demonstrate the potential of employing GPT-4 for order recovery, which has applications in scene understanding, image editing, autonomous vehicles, and robotics. 

\begin{table*}[!h]
\caption{Comparison between ground truth and GPT-4 predictions without category labels.}
\label{table:no_category}
\centering
\renewcommand{\arraystretch}{1.2}
\begin{tabular}{m{0.14\linewidth}
                |m{0.09\linewidth}
                |m{0.10\linewidth}
                |m{0.14\linewidth}
                |m{0.18\linewidth}}
\toprule
\multicolumn{1}{c|}{} 
& \multicolumn{2}{c|}{\textbf{Ground Truth}}
& \multicolumn{2}{c}{\textbf{GPT-4}} \\
\cmidrule(lr){2-3} \cmidrule(lr){4-5}
{\centering \textbf{Input Image}\par}
& \centering \textbf{Objects}
& \centering \textbf{Order Matrix}
& \centering\textbf{Detected Objects}
& \textbf{Predicted Order Matrix} \\
\midrule
    \includegraphics[width=\linewidth]{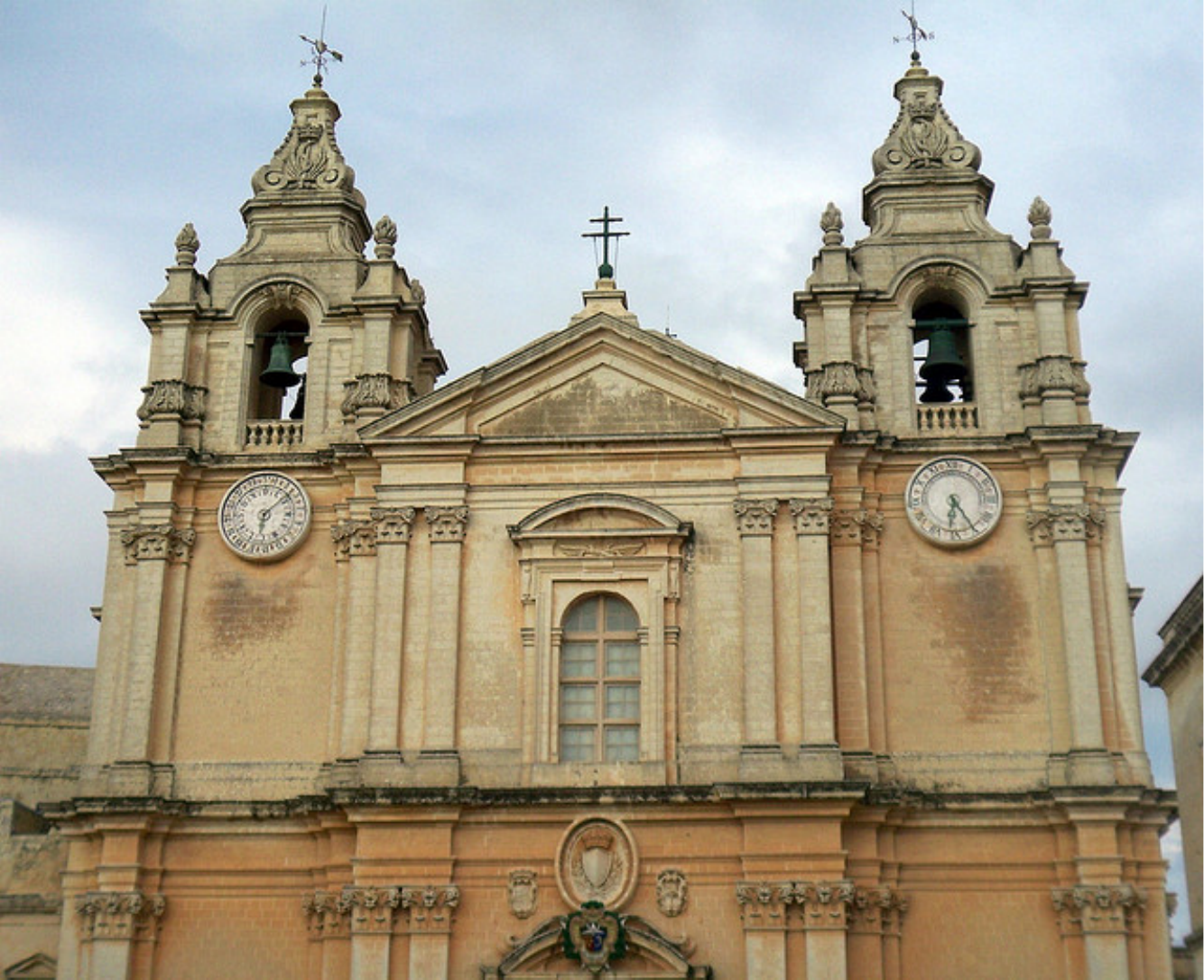}
& 0. Clock 0 \newline
1. Clock 1 \newline
2. Building 0

& \begin{verbatim}
[[ 0  0  1]
 [ 0  0  1]
 [-1 -1  0]]
\end{verbatim}

& 0. Clock on the right \newline
1. Cross \newline
2. Clock on the left \newline
3. Sky \newline
4. Pediment \newline
5. Building facade

& \begin{verbatim}
[[ 0  0  0  0  0  1]
 [ 0  0  0  1  0  0]
 [ 0  0  0  0  0  1]
 [ 0 -1  0  0  0  0]
 [ 0  0  0  0  0  1]
 [-1 0 -1 0 -1 0]]
\end{verbatim}

\\
\bottomrule
\end{tabular}
\end{table*}

\section{Conclusions}
Accurate object order recovery is fundamental for robust image and scene understanding. This work demonstrates the capability of GPT-4 in determining the occlusion relationships between objects within an image. Our method requires only the input image and a carefully engineered textual prompt to infer these relationships. Experimental results show that GPT-4 produces more accurate order predictions than existing baseline methods. This is due to the model's ability to leverage its extensive training on vast datasets, allowing it to discern complex visual patterns and apply rich semantic knowledge to deduce object occlusion order. This suggests that GPT-4 can be readily integrated into de-occlusion frameworks and contribute to other occlusion handling tasks. Future work could explore providing bounding box information to the model to ensure precise sequence alignment with ground truth labels, potentially leading to enhanced response accuracy.

\section*{Acknowledgment}
We would like to thank the ``Doctoral School of Applied Informatics and Applied Mathematics'' and the ``High Performance Computing Research Group'' of Obuda University for their valuable support. On behalf of the ``Development of Machine Learning Aided Metaheuristics'' project, we are grateful for the possibility to use HUN-REN Cloud \cite{heder2022past} which helped us achieve the results published in this paper.

\bibliographystyle{ieeetr}
\bibliography{references}

\end{document}